\documentclass[sigconf,natbib=true,nonacm]{acmart}
\AtBeginDocument{%
  }

\usepackage{booktabs}
\usepackage{enumitem}
\usepackage{amsmath}
\usepackage{stfloats}

\setlist{topsep=1pt, partopsep=0pt, parsep=0pt, itemsep=1pt}

\begin{document}

\title{LLM within MCP Matters: Measuring Inefficient Resource Utilization Driven by LLMs}
\titlenote{Accepted at the AgentSearch Workshop at SIGIR 2026, Melbourne, Australia (non-archival). This arXiv version is the authors' revision of the accepted extended abstract, incorporating reviewer feedback; the findings are unchanged.}

\author{Minhan Cho}
\authornote{Both authors contributed equally to this research.}
\email{zvezda@g.skku.edu}
\affiliation{%
  \institution{Remember \& Company AI Lab}
  \city{Seoul}
  \country{Republic of Korea}
}
\affiliation{%
  \institution{Sungkyunkwan University}
  \city{Seoul}
  \country{Republic of Korea}
}
\author{Soyoung Park}
\authornotemark[1]
\email{attorney.soyoung@gmail.com}
\affiliation{%
  \institution{National Assembly Research Service}
  \city{Seoul}
  \country{Republic of Korea}
}
\affiliation{%
  \institution{Sungkyunkwan University}
  \city{Seoul}
  \country{Republic of Korea}
}
\author{Kihyeon Jeong}
\email{jh.jeong@alphabridge.co.kr}
\affiliation{%
  \institution{AlphaBridge}
  \city{Seongnam-si}
  \state{Gyeonggi-do}
  \country{Republic of Korea}
}
\author{Byeongkyu Jeon}
\email{bk.jeon@alphabridge.co.kr}
\affiliation{%
  \institution{AlphaBridge}
  \city{Seongnam-si}
  \state{Gyeonggi-do}
  \country{Republic of Korea}
}
\author{Daejin Choi}
\authornote{Corresponding author.}
\email{djchoi@ewha.ac.kr}
\affiliation{%
  \institution{Ewha Womans University}
  \city{Seoul}
  \country{Republic of Korea}
}
\author{Jinyoung Han}
\authornotemark[2]
\email{jinyounghan@skku.edu}
\affiliation{%
  \institution{Sungkyunkwan University}
  \city{Seoul}
  \country{Republic of Korea}
}

\renewcommand{\shortauthors}{Cho et al.}

\begin{abstract}
  The Model Context Protocol (MCP) standardizes how servers expose data and tools to Large Language Models (LLMs).
A common server design embeds frequently used reference data, such as identifier lookup tables, directly in the \emph{server instructions}: the system-prompt text a server hands to the host application.
When a query concerns an entry of the embedded table, the model can act on it immediately instead of re-discovering the same information through a search tool.
We test whether client LLMs actually consume such instruction-embedded data, reporting a 54,000-trial study across 24~LLMs (9~Claude, 6~Gemini, 9~GPT) on a production legal-information MCP server.
A diagnostic condition that removes the competing search tool shows that failures are dominated by \emph{behavioral preference} rather than missing capability.
With search unavailable, 23 of 24~models read the embedded data reliably (hit ratio ${\geq}98\%$); with a search tool merely present, 9~models drop below 15\%.
A $2^3$ factorial analysis of three instruction-level interventions reveals strong interaction effects: combining all three restores ${\geq}86\%$ for 20 of 24~models, but individual interventions can backfire for specific model families.
Per-server prompt engineering is therefore a workaround rather than a fix; we argue that MCP host applications should provide an explicit mechanism that places server instructions ahead of tool selection in the client LLM's deliberation.
\end{abstract}

\ccsdesc[500]{Computing methodologies~Natural language processing}
\ccsdesc[500]{Computing methodologies~Intelligent agents}
\ccsdesc[500]{Computing methodologies~Multi-agent systems}

\keywords{Model Context Protocol, large language models, tool-augmented agents, instruction following}

\maketitle

\section{Background}

The Model Context Protocol (MCP)~\cite{anthropic2024mcp} standardizes how external data and tools are supplied to Large Language Models (LLMs), and it has been adopted rapidly across the ecosystem~\cite{ehtesham2025protocols}.
Under MCP, a host application (e.g., Claude Desktop or Cursor) connects to MCP servers, each of which exposes its capabilities through several channels; three are central to this study: (i)~\textbf{Resources}, structured data the host can fetch on demand; (ii)~\textbf{Tool schemas}, functions the host can execute through the function-calling API; and (iii)~\textbf{Server instructions}, system-prompt text in which the server describes how its resources and tools are meant to be used.
The host presents these capabilities, together with the user query, to an LLM, which we call the \emph{client LLM}; the client LLM plans the tool calls that the host then executes.
Embedding reference data in server instructions is an established design pattern: Block's Linear MCP server embeds its GraphQL schema in instructions\footnote{\url{https://engineering.block.xyz/blog/blocks-playbook-for-designing-mcp-servers}}, and legal-information servers embed identifier mappings for direct retrieval~\cite{lexlink2025}.

Despite this growing adoption, how effectively client LLMs exploit the capabilities a server exposes has been little explored.
Server instructions are the channel through which a server communicates its most efficient usage patterns, yet once a host has injected them into context, compliance rests on the client LLM's reasoning.
Prior evidence is not reassuring: MCPGauge reports low single-turn compliance with MCP server instructions~\cite{mcpgauge2025}, and instruction hierarchies fail under conflict even in frontier models~\cite{geng2026controlillusion}.

\section{Study Design}

\begin{table*}[t]
  \centering
  \scriptsize
  \caption{Hit ratio $\phi$ (\%) for the factorial and \texttt{no\_s} conditions (\textbf{bold}\,=\,100\%). Factors B, C, D defined in \S2. Each cell aggregates 250 trials (95\% binomial CI $\leq \pm 6.2$\,pp). $^\dagger$These four models scored exactly 100\% in every condition (250/250 trials per cell) and are collapsed into one row: cl-sonnet-4, cl-sonnet-4.5, cl-haiku-4.5, cl-opus-4.5.}
  \label{tab:embed}
  \vspace{-2pt}
    \centering
      \begin{tabular}{ll rrrrrrrrr}
        \toprule
        \textbf{Model} & \textbf{Date} & \textbf{base} & \textbf{B} & \textbf{C} & \textbf{D} & \textbf{BC} & \textbf{BD} & \textbf{CD} & \textbf{BCD} & \textbf{no\_s} \\
        \midrule
        \multicolumn{11}{l}{\textit{Anthropic Claude}} \\
        \midrule
        cl-3-haiku      & 2024-03 & 41  & \textbf{100} & \textbf{100} & 88  & \textbf{100} & \textbf{100} & \textbf{100} & \textbf{100} & \textbf{100} \\
        cl-s4,s4.5,h4.5,o4.5$^\dagger$ & -- & \textbf{100} & \textbf{100} & \textbf{100} & \textbf{100} & \textbf{100} & \textbf{100} & \textbf{100} & \textbf{100} & \textbf{100} \\
        cl-opus-4       & 2025-05 & 78  & \textbf{100} & \textbf{100} & \textbf{100} & \textbf{100} & \textbf{100} & \textbf{100} & \textbf{100} & \textbf{100} \\
        cl-opus-4.1     & 2025-08 & 77  & \textbf{100} & \textbf{100} & \textbf{100} & \textbf{100} & \textbf{100} & \textbf{100} & \textbf{100} & \textbf{100} \\
        cl-sonnet-4.6   & 2026-02 & 98  & \textbf{100} & \textbf{100} & 99  & \textbf{100} & \textbf{100} & \textbf{100} & \textbf{100} & \textbf{100} \\
        cl-opus-4.6     & 2026-02 & 63  & \textbf{100} & \textbf{100} & \textbf{100} & \textbf{100} & \textbf{100} & \textbf{100} & \textbf{100} & \textbf{100} \\
        \midrule
        \multicolumn{11}{l}{\textit{Google Gemini}} \\
        \midrule
        gem-2.0-flash   & 2025-02 & 17  & \textbf{100} & 91  & 81  & \textbf{100} & \textbf{100} & \textbf{100} & \textbf{100} & \textbf{100} \\
        gem-2.5-flash   & 2025-06 & 21  & 40  & 43  & 32  & 77  & 47  & 63  & 57  & 99  \\
        gem-2.5-pro     & 2025-03 & \textbf{100} & \textbf{100} & 96  & 98  & \textbf{100} & \textbf{100} & 98  & \textbf{100} & 99  \\
        gem-3-flash     & 2025-12 & 3   & 10  & 4   & 2   & 47  & 30  & 7   & 66  & \textbf{100} \\
        gem-3-pro       & 2025-11 & 2   & 27  & 60  & 7   & 91  & 89  & 85  & \textbf{100} & \textbf{100} \\
        gem-3.1-pro     & 2026-02 & 8   & 44  & 43  & 9   & 79  & 60  & 56  & 88  & \textbf{100} \\
        \midrule
        \multicolumn{11}{l}{\textit{OpenAI GPT}} \\
        \midrule
        gpt-4o-mini     & 2024-07 & 36  & \textbf{100} & 92  & 64  & \textbf{100} & \textbf{100} & 93  & \textbf{100} & \textbf{100} \\
        gpt-4.1-nano    & 2025-04 & 0   & 20  & 82  & 3   & \textbf{100} & 38  & 84  & \textbf{100} & 54  \\
        gpt-4.1-mini    & 2025-04 & 0   & 48  & 46  & 49  & 89  & 84  & 59  & 64  & \textbf{100} \\
        gpt-4.1         & 2025-04 & 99  & \textbf{100} & \textbf{100} & \textbf{100} & \textbf{100} & \textbf{100} & \textbf{100} & \textbf{100} & \textbf{100} \\
        gpt-5-nano      & 2025-08 & 1   & 98  & 9   & 26  & 99  & 99  & 36  & \textbf{100} & 99  \\
        gpt-5-mini      & 2025-08 & 0   & 66  & 0   & 0   & 77  & 76  & 2   & 86  & \textbf{100} \\
        gpt-5           & 2025-08 & 1   & 83  & 90  & 6   & \textbf{100} & 86  & 93  & \textbf{100} & \textbf{100} \\
        gpt-5.1         & 2025-11 & 65  & 98  & \textbf{100} & 82  & \textbf{100} & 91  & \textbf{100} & \textbf{100} & \textbf{100} \\
        gpt-5.2         & 2025-12 & 5   & 0   & \textbf{100} & 28  & \textbf{100} & 0   & 99  & 77  & \textbf{100} \\
        \bottomrule
      \end{tabular}
\end{table*}

We investigate whether, and by how much, the choice of client LLM changes how efficiently the same MCP server is used.
Our testbed is LexLink~\cite{lexlink2025}, a production MCP server that exposes 26~tools for Korean legal information retrieval over MCP protocol version \texttt{2025-06-18}.
Retrieving a law normally takes two steps: \texttt{eflaw\_search} maps a query to law identification numbers (IDs), and \texttt{eflaw\_service} fetches the content of one ID.
To make the common case cheaper, LexLink's server instructions embed a table of the 20 most frequently requested laws and their IDs (e.g., Civil Act $\to$ \texttt{001706}).
For these laws a single \texttt{eflaw\_service} call with the cached ID suffices (\texttt{eflaw\_service(id="001706")} for the Civil Act), saving both tokens and latency.
An efficient client LLM should therefore consult the embedded table first and fall back to \texttt{eflaw\_search} only when the query matches no listed law.\footnote{The experiment harness, server configurations, and per-trial data are available at \url{https://github.com/rabqatab/llm-in-mcp-matters}.}

We quantify this behavior with the \emph{hit ratio} $\phi(m,c)$: the fraction of trials, for model $m$ under condition $c$, in which the model's first tool call is \texttt{eflaw\_service} with the correct cached ID (exact match; any other first call, including a search call or a wrong ID, counts as a miss).

We evaluate 24~LLMs: 9~Claude, 6~Gemini, and 9~GPT models spanning multiple generations and capability tiers~\cite{schick2024toolformer,qin2024toolllm}.
All trials use single-turn prompting with automatic tool selection, temperature 1.0, and a 512-token output cap.
Temperature 1.0 is the default of all three provider APIs, so our results reflect out-of-the-box behavior rather than a tuned setting; defaults are not guaranteed to induce comparable sampling across vendors, so cross-provider gaps should be read qualitatively.
Five user queries each ask for the first article of a law listed in the embedded table, at three levels of ID-matching difficulty: short exact names (Q1--Q2), longer formal titles (Q3--Q4), and an abbreviation (Q5).
With 24~models $\times$ 9~conditions $\times$ 5~queries $\times$ 50~rounds, the study comprises 54,000~trials\footnote{The workshop version reported 72,000 trials; that count included three resource-channel conditions that are not part of the analysis reported here.
All results below are unchanged.}.
Each cell of Table~\ref{tab:embed} pools 250~trials (five queries $\times$ 50~rounds); under an independence assumption, the half-width of a 95\% binomial confidence interval is at most $\pm$6.2 percentage points (pp).
Trials cluster within the five fixed queries, however, so cell values are averages over this query set rather than estimates for arbitrary queries (\S3 shows per-query decompositions).

Each model runs under 9~conditions built from three instruction-level interventions:
\begin{itemize}[nosep, leftmargin=*]
  \item \textbf{B} (\emph{directive}): the base setup augmented with a \texttt{CRITICAL} header before the law-ID table (``\emph{For any law listed in the table below, you MUST call \texttt{eflaw\_service(id=...)} directly using the cached ID. DO NOT call \texttt{eflaw\_search} for these laws}'').
  \item \textbf{C} (\emph{few-shot example}): the base setup augmented with a worked example contrasting correct and incorrect tool-call patterns.
  \item \textbf{D} (\emph{tool-description hint}): the base setup with \texttt{eflaw\_search}'s description modified to carry a self-referential warning (``\emph{Before using this tool, check the ID table in your instructions; if the law is listed there, use \texttt{eflaw\_service(id=...)} directly instead}'').
\end{itemize}
The $2^3=8$ combinations (\texttt{base}, \texttt{B}, \texttt{C}, \texttt{D}, \texttt{BC}, \texttt{BD}, \texttt{CD}, \texttt{BCD}) form a full factorial.
A ninth condition, \texttt{no\_s}, removes \texttt{eflaw\_search} entirely, leaving the embedded table as the only source of IDs.
It is a diagnostic: a model that succeeds under \texttt{no\_s} but fails at baseline does not lack the ability to use the embedded data.
Removing the tool also shrinks the action space, so we read \texttt{no\_s} as a capability control rather than a pure preference probe; what it establishes is that the baseline failure is not a parsing limitation.

\section{Results}

The \texttt{no\_s} diagnostic provides a sharp test (Table~\ref{tab:embed}).
With the search tool removed, 23 of 24~models use the embedded data in at least 98\% of trials (22 of them at ${\geq}99\%$), so nearly every model \emph{can} consume instruction-embedded data.
With the search tool present, baseline $\phi$ spans 0--100\%, and 9~models fall below 15\%.
In this setup, the mere presence of a search tool in the tool list is thus sufficient to override instruction-embedded data, even though the data makes searching redundant.
We call this failure mode a \emph{behavioral preference}, in contrast to a capability limitation.\footnote{Success under \texttt{no\_s} reflects reading the table, not parametric memory of the IDs: in one of the omitted resource-channel conditions, which removes the embedded table while keeping tools available, no model produces a correct cached ID in more than 10\% of trials.} Prior work documented biased tool selection~\cite{faghih2025toolpref,blankenstein2026biasbusters} and limited instruction compliance~\cite{mcpgauge2025,agentif2025}; our diagnostic isolates tool availability as the trigger in this setting.
Whether the preference reflects trained bias or a rational taste for verifiable tool output, the practical consequence for resource utilization is the same.
The one exception, gpt-4.1-nano (54\% under \texttt{no\_s}), shows that a genuine capability deficit also exists at the low end: the weakest model in our pool often fails to match a query to the table even when no alternative tool exists, so its baseline failure mixes both components.
Aggregate cells can likewise mask query-difficulty effects: cl-opus-4.6's 63\% baseline decomposes into 100\% on the three easier queries but 16\% and 0\% on the two hardest, and gpt-4.1-nano's \texttt{no\_s} failures concentrate on the longer formal titles (Q3--Q4).

$\phi$ is also non-monotonic across model generations: GPT-4.1 (99\%) drops to GPT-5 (1\%), Gemini 2.5-pro (100\%) to 3-pro (2\%), and Claude Opus 4.5 (100\%) to Opus 4.6 (63\%).
Model recency does not predict efficient resource utilization.
We can only speculate about the mechanism.
Post-training that rewards successful tool calls on agentic benchmarks would teach exactly the observed policy (when a matching tool exists, call it), and provider differences in how heavily system-prompt content is weighted during that training would explain why the regression is family-specific.
Separating these hypotheses requires training details that providers do not publish.

The $2^3$ factorial analysis tests whether instruction-level interventions mitigate the bypass.
B and C have similar aggregate main effects ($+$19.9 and $+$19.8\,pp averaged across all 24~models) but high cross-model variance ($\text{SD}{\approx}24$\,pp), because they act on different model profiles: B dominates for GPT-5-nano ($+$80.9\,pp), C for GPT-5.2 ($+$85.6\,pp).
Both aggregate effects are nonzero at $p{<}0.001$ by one-sample $t$-tests over the 24 model-level effects, which we read descriptively since models cluster by provider.
Individual factors can also fail outright or reverse: GPT-5.2 drops from 5\% to 0\% under B alone ($z{=}3.65$, $p{<}0.001$), yet recovers to 100\% under BC, an interaction that a one-factor-at-a-time design would never reveal.
Prior work noted over-prompting risks~\cite{fewshotdilemma2025}; the factorial pinpoints which combinations backfire for which families.
Combining all three factors (\texttt{BCD}) achieves $\phi \geq 86\%$ for 20 of 24~models.

\section{Concluding Remarks}

Across 24~LLMs and 9~conditions, client LLMs systematically favor invoking a search tool over consuming equivalent data already present in server instructions.
Nearly all models can read the embedded data ($\phi \geq 98\%$ under \texttt{no\_s} for 23/24), yet 9~models fall below 15\% once the search tool is available; instruction-level interventions recover much of the gap (\texttt{BCD} ${\geq}86\%$ for 20/24) while single factors can backfire (GPT-5.2: 5\%\,$\to$\,0\% under B alone).
All of this evidence comes from one server and one domain.
An expanded study, to be reported separately, repeats the design on four additional purpose-built MCP servers spanning geographic, academic, database, and messaging tasks, and evaluates a host-side mitigation.
Its preliminary results are consistent with the pattern reported here, but we cite them as an outlook, not as evidence.

Three implications follow.
First, capability assessments of MCP servers should be reported jointly with the client environment, since a server's effective capability depends on how it exposes data and on which LLM consumes it.
Second, per-server prompt engineering is brittle: a directive that rescues one model breaks another.
We therefore argue that host applications should provide an explicit mechanism placing server instructions early in the client LLM's deliberation, for example by pre-injecting them as a system-priority directive or by enforcing a consult-instructions-first check before each tool selection.
Evaluating such mechanisms is part of the expanded study.
Third, MCP server developers can act today.
Include few-shot examples, whose main effect was positive or negligible for every model in our grid, whereas a directive alone can backfire.
Combine B, C, and D for the broadest coverage.
Validate against the weakest client models the server is expected to serve.

Several limitations bound these findings.
All evidence derives from a single server and domain, as noted above.
Our metric captures first-call efficiency rather than end-to-end task success; because a correct first call retrieves the intended law directly, the two coincide on this testbed, but we do not score the final natural-language answer.
All trials are single-turn, whereas a multi-turn agent may self-correct at additional token and latency cost.
Each intervention is one concrete wording, so measured effects attach to B, C, and D as instantiated here rather than to their categories.
Finally, our conclusions concern lookup-style workflows, the regime that instruction-embedded data is designed to serve; tasks where searching is genuinely informative fall outside this study's scope.

\begin{acks}
This work was supported by the IITP(Institute of Information \& Communications Technology Planning \& Evaluation)-ITRC(Information Technology Research Center) grant funded by the Korea government(Ministry of Science and ICT)(IITP-2026-RS-2024-00436936). This research was supported by the MSIT(Ministry of Science and ICT), Korea, under the ICAN(ICT Challenge and Advanced Network of HRD) support program(IITP-2026-RS-2023-00259497) supervised by the IITP(Institute for Information \& Communications Technology Planning \& Evaluation). This research was supported by the MSIT(Ministry of Science and ICT), Korea, under the Graduate School of Virtual Convergence support program(IITP-2026-RS-2023-00254129) supervised by the IITP(Institute for Information \& Communications Technology Planning \& Evaluation).
\end{acks}


\bibliographystyle{ACM-Reference-Format}
\bibliography{main}

\end{document}